\documentclass[10pt,journal,compsoc]{IEEEtran}
\ifCLASSOPTIONcompsoc
\usepackage[nocompress]{cite}
\else
\usepackage{cite}
\fi

\usepackage{times}
\usepackage{epsfig}
\usepackage{graphicx}
\usepackage{amsmath}
\usepackage{amssymb}
\usepackage{times}
\usepackage{epsfig}
\usepackage{graphicx}
\usepackage{amsmath}
\usepackage{amssymb}
\usepackage{multirow}
\usepackage{url}
\usepackage{caption}
\usepackage{subfigure}
\usepackage{float} 
\usepackage{algpseudocode,algorithm}
\usepackage{gensymb}
\usepackage{booktabs}
\usepackage{xcolor}
\usepackage{tablefootnote}
\usepackage{threeparttable}

\graphicspath{{figure/}}

\usepackage[breaklinks=true,bookmarks=false]{hyperref}

\ifCLASSINFOpdf
\else
\fi
\begin{document}

\title{CoSeg: Cognitively Inspired Unsupervised Generic Event Segmentation }

\author{Xiao Wang,
	Jingen Liu,~\IEEEmembership{Senior Member,~IEEE,}
	Tao Mei, ~\IEEEmembership{Fellow,~IEEE,}
	and~Jiebo Luo,~\IEEEmembership{Fellow,~IEEE}
	\IEEEcompsocitemizethanks{
		\IEEEcompsocthanksitem Xiao Wang is with the Department
		of Computer Science, Purdue University, West Lafayette, IN, 47906, USA.
		\IEEEcompsocthanksitem Jingen Liu is with JD AI Research,  
		Mountain View, CA, USA, 94039. Email: jingen.liu@gmail.com
		\IEEEcompsocthanksitem Tao Mei is with JD AI Research, Beijing, China. 
		\IEEEcompsocthanksitem Jiebo Luo is with University of Rochester, Rochester, 14627, USA.
	}	
	
	\thanks{Manuscript received Sep 10, 2021; revised xx,xx, accepted xx,xx.}
}

\markboth{IEEE Transactions on Neural Networks and Learning Systems,~Vol.~x, No.~x, xx~2021}%
{Wang \MakeLowercase{\textit{et al.}}: CoSeg: Cognitively Inspired Unsupervised Generic Event Segmentation }

\IEEEtitleabstractindextext{%
	\begin{abstract}
		Some cognitive research has discovered that humans accomplish event segmentation as a side effect of event anticipation. Inspired by this discovery, we propose a simple yet effective end-to-end self-supervised learning framework for event segmentation/boundary detection. Unlike the mainstream clustering-based methods, our framework exploits a transformer-based feature reconstruction scheme to detect event boundary by reconstruction errors. This is consistent with the fact that humans spot new events by leveraging the deviation between their prediction and what is actually perceived. Thanks to their heterogeneity in semantics, the frames at boundaries are difficult to be reconstructed (generally with large reconstruction errors), which is favorable for event boundary detection. Additionally, since the reconstruction occurs on the semantic feature level instead of pixel level, we develop a temporal contrastive feature embedding module to learn the semantic visual representation for frame feature reconstruction. This procedure is like humans building up experiences with ``long-term memory''. The goal of our work is to segment generic events rather than localize some specific ones. We focus on achieving accurate event boundaries. As a result, we adopt F1 score (Precision/Recall) as our primary evaluation metric for a fair comparison with previous approaches. Meanwhile, we also calculate the conventional frame-based MoF and IoU metric. We thoroughly benchmark our work on four publicly available datasets and demonstrate much better results. 

	\end{abstract}
	
}

\maketitle

\section{Introduction}
\label{sec:introduction}
Temporal event segmentation in untrimmed long videos is crucial to complex activity understanding and has received increasing attention recently. Here, an event is defined as ``a meaningful video segment of time at a given location that an observer conceives to have a beginning and an end''~\cite{zacks2001event}, and a complex activity generally consists of a sequence of such events. 

To automatically segment an activity into events, previous supervised approaches~\cite{ding2018weakly,kuehne2014language,kuehne2016end,lea2016segmental,richard2017weakly} are trained with dense frame-level annotations. Although they have achieved good performance, the expensive and time-consuming annotation is a hindrance to large-scale applications. In contrast, weakly supervised methods~\cite{alayrac2016unsupervised,laptev2008learning,malmaud2015s,bojanowski2014weakly,ding2018weakly} attempt to acquire supervision signals from the video's side information including subtitles, text scripts, and audios. However, such side information may not be reliable for the training due to the lack of precise alignment with the video in temporal order. 

\begin{figure}[t!]
\centering
%
\includegraphics[width=\linewidth]{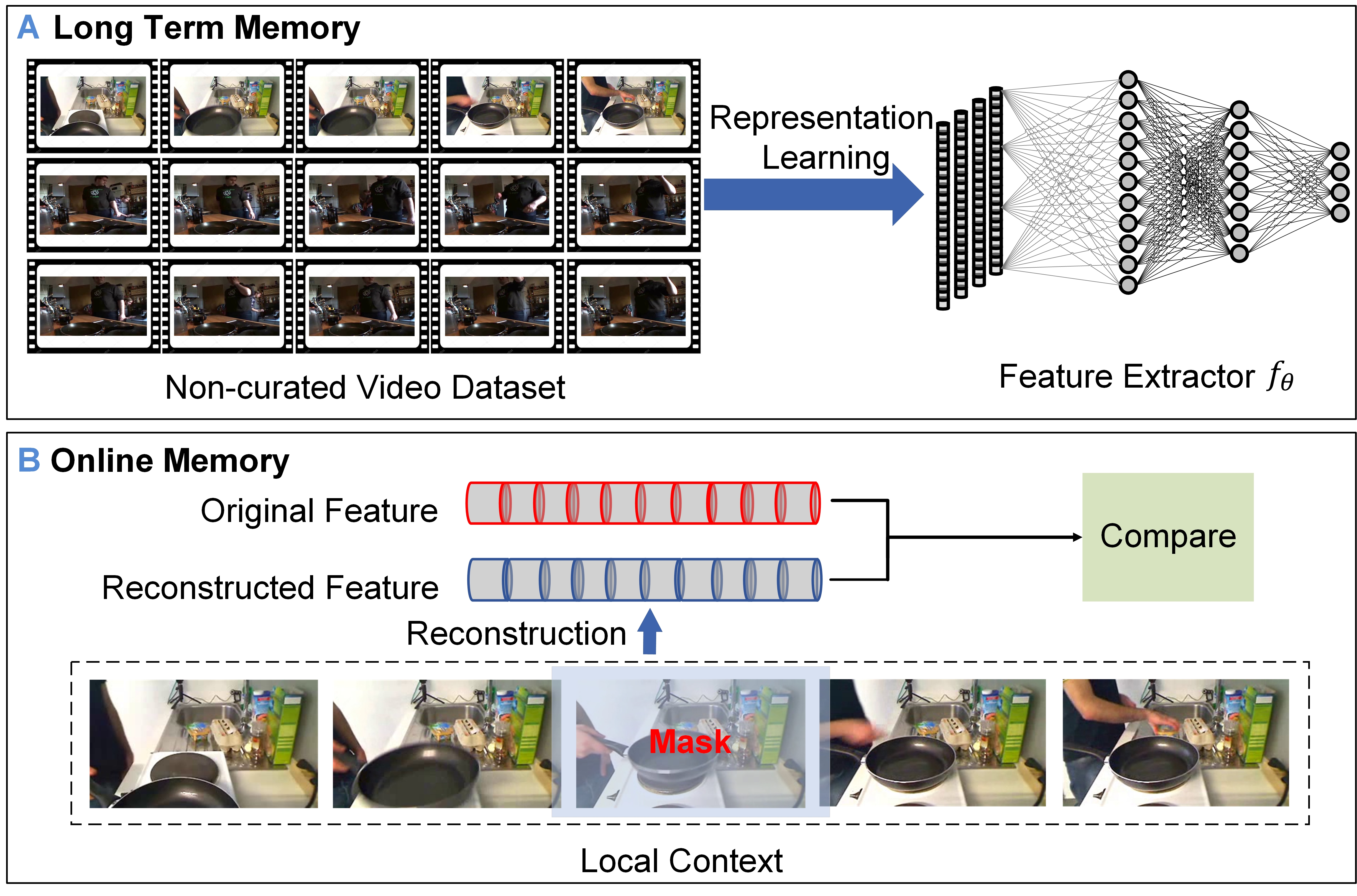}
\caption{The motivation of our CoSeg event segmentation. (a) To imitate human's long-term memory, a self-supervised feature representation network is trained to distinguish intra-video frames from inter-video frames in the video dataset. (b) To mimic the online memory, a frame feature reconstruction network is trained to capture local temporal relationships within a local snippet context. 
}
\label{fig:memory}
\end{figure}

Therefore, unsupervised event segmentation is receiving increasing attention. Based on the observation that different frames of the same event usually share similar visual features, most recent unsupervised approaches~\cite{bhatnagar2017unsupervised,kukleva2019unsupervised,lea2016segmental,sener2018unsupervised,vidalmata2020joint} exploit various clustering mechanisms to uncover the sub-actions/events in the dataset and then apply the clusters for event segmentation. However, the clustering procedure disregards important and meaningful temporal relationships between video segments. Besides, it is difficult to determine the number of clusters when conducting clustering on video datasets, especially for an uncurated one. As a result, the clustering-based approaches are not good at dealing with generic event segmentation for videos ``in the wild''. Essentially, the clustering mechanism is not 
well aligned with humans perception system, because humans perceive an event boundary without explicitly grouping sub-actions. 

In fact, some recent cognitive studies ~\cite{kurby2008segmentation,zacks2007event,zacks2006event} have illustrated humans segment events automatically and subconsciously. According to the Event Segmentation Theory (EST)~\cite{zacks2007event}, the human perceptual system spontaneously segments an activity into different events as a side effect of anticipating upcoming events. In other words, a new event is triggered by the human perceptual system due to the failure of event anticipation or reconstruction. To further understand this phenomenon, additional cognitive experiments~\cite{zacks2006event} have been conducted to demonstrate that the event anticipation procedure is closely related to the working memory system, which includes the \textit{long-term memory} linked to previously learned knowledge and the \textit{online memory} associated with the present activity. Humans exploit the working memory to anticipate the upcoming stream and its failure to identify the event boundary.    

Inspired by these cognitive studies, in this paper we propose an end-to-end self-supervised generic event segmentation framework, called CoSeg. As shown in Fig.~\ref{fig:memory}, we devise a transformer-based frame feature reconstruction scheme to identify the event boundary. It effectively exploits the temporal relationships of video frames to generate an arbitrary frame in its feature space. As we observed, the frames around an event boundary are usually semantically heterogeneous. As a result, it is difficult to reconstruct those boundary frames, and such reconstructions generally carry higher reconstruction errors, which can be exploited to detect the boundary.

Our frame reconstruction occurs in the feature space instead of the pixel space, because our system aims at ``understanding'' a frame to tell if it belongs to the current event. To better support the reconstruction in feature space, we develop a self-supervised feature representation learning network, which utilizes a contrastive learning mechanism to distinguish if two frames from a typical local context / event. Leveraging the carefully designed temporal positive and negative training pairs, our Temporal Contrastive Feature Embedding (TCFE) learns a solid semantic feature representation for frame reconstruction. This feature representation learning acts like the mechanism when humans build up their long-term memory.  

Unlike the mainstream clustering-based unsupervised methods, our framework opens a new avenue for unsupervised generic event segmentation. It can be trained with any non-curated dataset (no pre-defined action is needed), because our approach does not rely on the discovery of explicit common patterns (i.e., action categories or pseudo labels). As shown in Table \ref{tab:transfer}, a model that is trained on a different dataset can still achieve comparable results, which proves the transferability of our approach. Thanks to the contrastive feature embedding, our model is more generalized to avoid over-segmentation problem. In addition, our reconstruction can be conducted within a short range of frames (i.e., local context), which acts like humans' ``online memory''.   

In addition, in contrast to some conventional works in event segmentation and action detection which focus on localizing pre-defined action classes~\cite{bhatnagar2017unsupervised,sener2018unsupervised,kukleva2019unsupervised,bhatnagar2017unsupervised,vidalmata2020joint}, our CoSeg system aims at detecting generic and taxonomy-free event boundaries without pre-defined action categories. Specific action localization is out of the scope of our work. Our work can be categorized into the Generic Event Boundary Detection (GEBD) \cite{shou2021generic}, which detects generic event boundaries such that a video is divided into natural, meaningful segments. Compared to class-specific event segmentation, the GEBD task is more general and applicable to train and evaluate on long-frame datasets, such as TAPOS~\cite{shao2020intra} and Kinetics~\cite{kay2017kinetics}. It is valuable to various applications including video summarization and video-level classification \cite{shou2021generic}.

In short, our contributions can be summarized as follows: 
\begin{itemize}
    \item To the best of our knowledge, we are the first to leverage cognitive studies on how humans segment events to design a simple yet effective end-to-end self-supervised framework for generic event segmentation.

    \item Unlike previous clustering-based approaches, our CoSeg system can work on non-curated datasets. It is able to conduct generic event segmentation, not only limited to predefined events. Hence, it can scale to any generic videos.
    
   \item Compared with previous methods, our CoSeg has demonstrated strong model transferability across different datasets and better generalization. As a result, it can alleviate the over-segmentation issue and the problem of overfitting to dominant classes.
   
   \item Our CoSeg has been extensively evaluated on four widely used benchmark datasets: Kinetics-GEBD dataset~\cite{shou2021generic}, Breakfast ~\cite{kuehne2014language}, 50 Salads~\cite{alayrac2016unsupervised}, and INRIA Instructional ~\cite{stein2013combining}. All experiments have demonstrated that our CoSeg can outperform previous approaches by a large margin in terms of various metrics.  
    
\end{itemize}

\section{Related Work}

\begin{figure*}[!htb]

\centering
%
\includegraphics[width=0.9\linewidth]{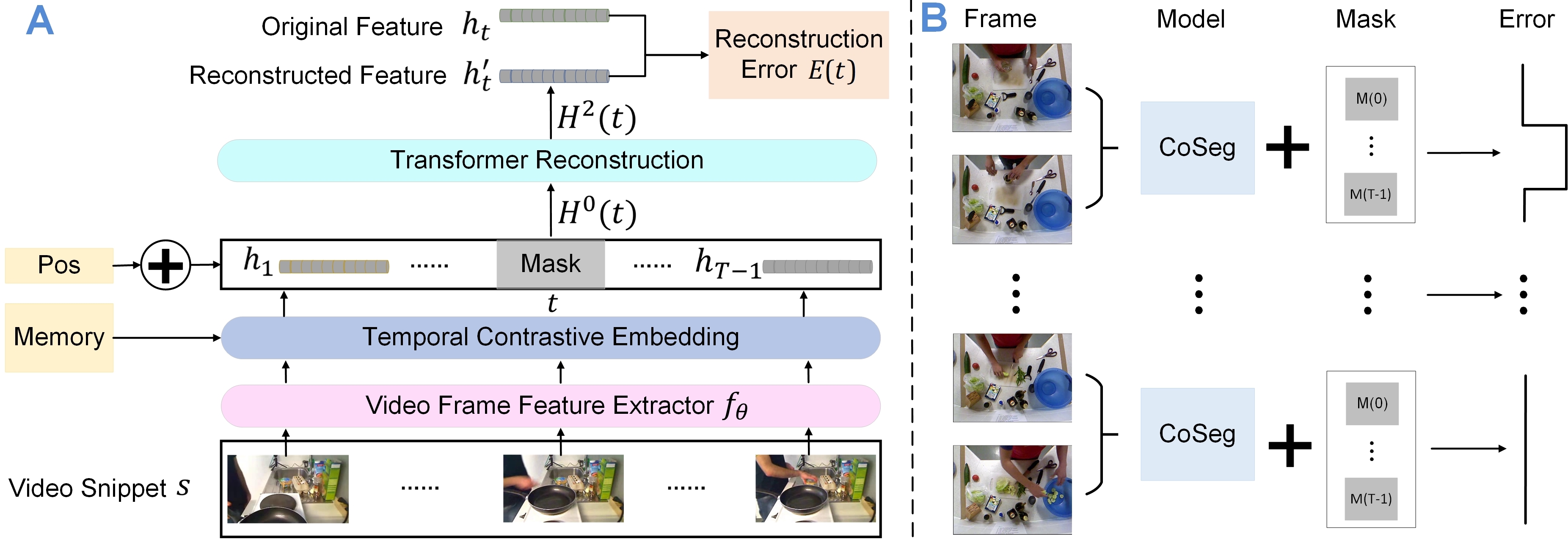}
\caption{\textbf{Overall framework of our CoSeg. } The end-to-end event segmentation system consists of two modules: Contrastive Temporal Feature Embedding~(CTFE) and temporal Frame Feature Reconstruction (FFR). \textbf{(A)} The training procedure of CoSeg. \textbf{(B)}. The inference procedure of CoSeg.}
\label{fig:framework}
\end{figure*}



\noindent\textbf{Supervised Event Segmentation.} The fully supervised approaches leverage densely annotated frame-level annotations to train classifiers and then assign event labels to each frame. The event segmentation is conducted by temporally grouping the frames with the same labels into some coherent ``chunks''. The representative frame-wise labeling approaches include support vector machine~(SVM)~\cite{kuehne2014language}, Hidden Markov Models~(HMM)~\cite{kuehne2014language}, Spatio-temporal Convolutional Neural networks~(SCN)~\cite{lea2016segmental} and Temporal Convolutional Neural networks~(TCN)~\cite{lea2017temporal}. 

\noindent\textbf{Weakly Supervised Event Segmentation.} To alleviate the need for a large amount of annotations, researchers have explored the weakly supervised methods, which attempt to train a model with less supervision. Some works~\cite{alayrac2016unsupervised,laptev2008learning,malmaud2015s,bojanowski2014weakly,ding2018weakly} propose to leverage the video's additional modalities, such as subtitles, text scripts, and audios, to provide supervision signal for the model training. One of their assumptions is acceptable alignment between visual frames and other modalities. Unfortunately, most time the quality of the alignment is not guaranteed. To overcome this limitation, some works ~\cite{huang2016connectionist,kuehne2017weakly,ding2018weakly,richard2017weakly,richard2018neuralnetwork} try to use the temporal order of actions in a video without alignment for weak supervision. 

\noindent\textbf{Unsupervised Event Segmentation.} Video segment clustering is the most widely used unsupervised mechanism by previous work~\cite{bhatnagar2017unsupervised,sener2018unsupervised,kukleva2019unsupervised,bhatnagar2017unsupervised,vidalmata2020joint}. The basic idea is grouping frames into sub-actions or events, which results in event segmentation. Leveraging the advanced frame representation by the pre-trained network on ImageNet, researchers have proposed various optimization methods to achieve better temporal clusters, such as Frank-Wolfe algorithm~\cite{bojanowski2017unsupervised}, Generalized Mallows Model~(GMM)~\cite{sener2018unsupervised}, Viterbi algorithm~\cite{kukleva2019unsupervised}, and U-Net decoding~\cite{vidalmata2020joint}. Although the clustering procedure is similar to the aforementioned long-term memory, the clustering results (i.e., segmentation) is susceptible to the number of clusters, which is hard to determine. In addition, clustering ignores the temporal order of events, which is an essential clue for event segmentation. 


Cognitive research \cite{zacks2001event,zacks2006event,zacks2007event,kurby2008segmentation} shows that human segmenting events is a side effect of human anticipation. Thus, Aarkur et al.~\cite{aakur2019perceptual} propose a RNN (i.e., LSTM) framework to predict the incoming frame using the past information and utilize the prediction errors to segment events. However, we argue that people generally rely on not only the past but also the temporally local context to tell the change of events. Therefore, we formulate event segmentation as a reconstruction problem rather than a predictive model. Most importantly, our novel end-to-end model can mimic human's working memory~\cite{zacks2007event}, which is associated with human's capability of event anticipation according to studies in \cite{zacks2007event,zacks2006event}. We use contrastive learning to train the task-related visual feature representation by distinguishing the intra-video and inter-video frames, which is an analogy for human's long-term memory learning. Meanwhile, we adopt the bidirectional transformer architecture to reconstruct masked frames in the feature space, imitating humans' online memory in the event segmentation. Our model significantly outperforms the one-directional predictive model. The recent U-Net-based work ~\cite{vidalmata2020joint} also exploits temporal frame reconstruction to learn feature representation, but it focuses on the pixel level reconstruction and turns to traditional unsupervised clustering for event segmentation.   


\begin{figure*}[]
\centering
%
\includegraphics[width=\linewidth]{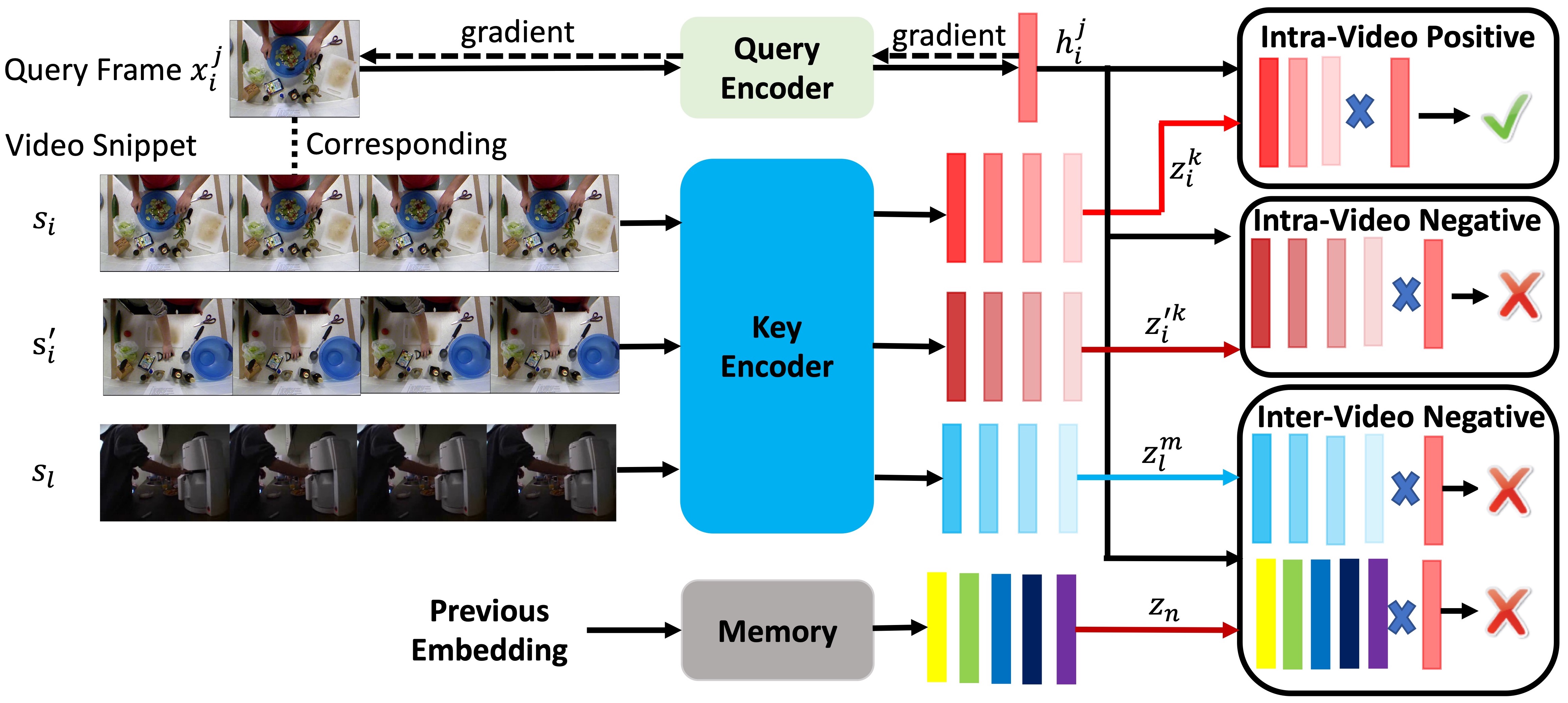}
\caption{ \textbf{The illustration of positive/negative pairs formation for contrastive embedding.} Given a set of video segments, e.g., snippet $s_i$ and $s'_i$ from a same video and snippet $s_l$ from a different video, let $x_{i}^{j}$ be a query frame of $s_i$ and $h_i^{j}$ be the corresponding query code. The other sampled frames are projected by the key encoder. There could be three types of pairs: the intra-segment positives from the same segment $s_i$, the intra-video negatives from a same video but different segment $s'_i$, the inter-video negatives from different segments $s_l$ or the memory carrying previous embedded frames.}
\label{fig:contrastive}
\end{figure*}

\noindent\textbf{Self-supervised learning.} has been widely explored recently ~\cite{he2020momentum,chen2020simple,caron2020unsupervised,wang2019enaet}. Among different self-supervised learning methods~\cite{bojanowski2017unsupervised,he2020momentum,donahue2016adversarial,donahue2019large,asano2019self,caron2018deep}, contrastive learning~\cite{hadsell2006dimensionality,wu2018unsupervised,chen2020simple,he2020momentum,grill2020bootstrap,hu2020adco}, which tries to maximize the agreement of positive pairs and disagreement of negative pairs, obtains better performance. The design of positive and negative pairs is critical. Meanwhile, similar contrastive idea is also adopted in video tasks~\cite{han2020self,fayyaz2020sct,han2020memory}. \cite{han2020self} designs the intra-frame and inter-frame to boost the performance on action recognition tasks. \cite{fayyaz2020sct} learns video representation by combining contrastive learning in RGB and flow feature space.  
Compared to them, we design temporal positive/negative pairs specifically for event boundary detection. Our feature reconstruction is also inspired by another popular self-supervised approach with ``masked token reconstruction'', which has been very successful in many applications~\cite{vaswani2017attention,devlin2018bert,brown2020language}. 
\section{Proposed End-to-End Method}
\label{sec:methods}

Given a generic video, the event segmentation task aims at detecting the boundary (frames) of two events. To this end, 
most unsupervised event segmentation approaches usually carry out frame clustering over a curated video dataset to label every frame with a pseudo-event label (i.e., cluster ID). However, clustering the high-dimensional frame features is very challenging, and its performance is sensitive to the number of clusters that is unknown for non-curated data. In fact, we humans segment events by detecting the deviation between what we are anticipating/expecting and perceiving. Inspired by this, in this section, we present a novel unsupervised event segmentation framework, which leverages the frame feature reconstruction errors to spot the event boundary.          

Fig.~\ref{fig:framework} shows the overview of our end-to-end unsupervised learning system (CoSeg), which consists of two major modules: Contrastive Temporal Feature Embedding (CTFE) and Frame Feature Reconstruction (FFR). Given a video segment/snippet $s$, which is a sequence of $T$ frames, we first apply feature extractor $f_\theta$ to obtain $T$ features $h_{t}(t\in [0, T-1])$. During training, the CTFE module is framed as self-supervised contrastive feature learning, taking intra-segment frames as positive pairs and inter-segment frames as negative pairs. The goal is to learn a distinctive semantic representation to distinguish event frames, acting like humans' ``long-term memory''. Meanwhile, we modify the transformer encoder to train the FFR module such that it can successfully reconstruct the masked frame feature $h_{t}$ of any given video snippet. During inference, our system tries to use FFR to reconstruct a frame feature as $h_{t}'$, and the reconstruction error $h_{t} $- $h_{t}'$ is exploited to tell if it is a boundary frame. This process is similar to that of humans, as humans are good at anticipating intra-event content rather than the transitions of two events. Likewise, the FFR module is usually worse at reconstructing event boundary frames due to its heterogeneous local context. 

During the learning process, the objective function of our CoSeg is to minimize the following loss,
\begin{equation}
\label{eq:overallloss}
\centering
 \mathcal L=\mathcal L_{C} + \beta*\mathcal L_R,
\end{equation}
where $\mathcal L_{C}$ is the loss for contrastive temporal feature embedding (CTFE, see Section \ref{sec:contrastive}), and $\mathcal L_R$ is the loss applied to the frame feature reconstruction (see Section \ref{sec:reconstruction}).

\subsection{Contrastive Temporal Feature Embedding}
\label{sec:contrastive}
In general, a video event consists of a sequence of semantically correlated frames. Namely, the neighbor frames are more likely to be semantically similar than frames sampled at long time intervals. In line with this observation, we propose a contrastive temporal feature embedding scheme to learn a discriminative frame representation. Essentially, it projects the semantically similar frames closer, while pushes away the dissimilar ones. Leveraging this learning by comparison, our framework transforms frames into a new representation being more semantically distinguishable. As illustrated in Fig.~\ref{fig:contrastive}, the positive pairs of contrastive learning consist of intra-segment frames, and the negative pairs come from inter-segment frames of other snippets from the same or other videos, or frames in the memory.  

More specifically, we select a batch of $B$ videos from a non-curated video dataset, and then we randomly extract $X$ non-overlapped video snippets of $T$ frames from each video. It results in $L=B*X$ video snippets in total. Let $x_{i}^{j}$, the $j$-th frame of snippet $s_{i}$, be a query frame. We select the neighbor frames of query $x_{i}^{j}$ in $s_{i}$ to form positive pairs. This is based on the assumption of the temporal semantic homogeneity of an event. Next, we form three types of negative pairs associated with query $x_{i}^{j}$: 1) intra-video negative pairs: the negative frame comes from the same video as $x_{i}^{j}$ but from a different snippet. 2) inter-video negative pairs: the negative frame is selected from any snippet $s_l$ extracted from a different video of $s_i$. 3) memory negative pairs: the negative frame comes from frames embedded in the memory during previous iterations. Please note, we only store one embedded frame per video segment in the memory, a FIFO queue with the size of $K$. This design can greatly increase the number of representative frames to boost its performance. During training, our goal is to increase the similarity of the embedding of positive pairs but decrease that of negative pairs by minimizing the following loss function:  
\begin{equation*}
\label{eq:contrastive0}
\centering
\mathcal L_C = \mathbb{E}_{i\in[0,L-1], j \in [0,T-1]}
\left[-\frac{1}{T-1}\sum_{k=0,k\ne j}^{k=T-1}\log P(i,j,k)\right]
\end{equation*}

with,
\begin{equation*}
\label{eq:pair}
\centering
\left\{  
\begin{array}{l} 
P(i,j,k)= \frac{Q^+(i,j,k)}{Q^+(i,j,k) + Q_{1}^-(i,j)+Q_{2}^-(i,j) }\\
Q^+(i,j,k)=\exp(sim(h_{i}^j,z_{i}^k)/\tau)\\
Q_{1}^-(i,j) = \sum_{l=0,l\ne i}^{L-1}\sum_{m=0}^{T-1}\exp(sim(h_{i}^j,z_{l}^{m})/\tau)\\
Q_{2}^-(i,j)=\sum_{n=0}^{K-1}\exp(sim(h_{i}^j,z_{n})/\tau)\\
\end{array}  
\right. 
\end{equation*}
where $sim$ denotes the cosine similarity; $Q^+(i,j,k)$, $Q_1^{-}(i,j)$, $Q_2^{-}(i,j)$ are the exponential temperature smoothed similarities of positive pairs, and the sum of exponential temperature smoothed similarities of inter-video and intra-video negative pairs and negative pairs from memory, respectively; $h$ and $z$ are the corresponding feature by query encoder $f_\theta$ and key encoder $g_\theta$, respectively; $\tau$ is the temperature set to 0.2, and $L$ is total number of video snippets in the batch. Here the query and key encoder architecture is widely used in contrastive learning, please refer to \cite{he2020momentum} for details.

\subsection{Frame Feature Reconstruction}
\label{sec:reconstruction}

As we know, the transition frames between video events are usually inconsistent and thus less predictable. Accordingly, we develop an unsupervised feature reconstruction approach to detect these event boundaries, because we conjecture that boundary frames usually carry higher reconstruction errors than non-boundary ones. Unlike previous pixel-level image reconstruction~\cite{vidalmata2020joint}, however, our frame reconstruction is conducted in a high-level semantic feature space. Namely, our approach aims at reconstructing a frame's semantic representation trained by CTFE (section ~\ref{sec:contrastive}).          


\noindent\textbf{Module Input.} Let us assume $x_t$ is the masked frame to be reconstructed from its local context of segment $s$ with $T$ frames. By applying feature extractor $f_\theta$ to its frames, we obtain the video segment's semantic representation as $H=\{h_{0} , ..., h_{T-1}\}$. When training the reconstruction network, $h_t$ is a masked element, while during inference it should be reconstructed from its local context. In addition, we apply position embedding to all frames of $s$ as $E_{pos}\in \mathbb{R}^{T\times D}=\{pos_{0},... pos_{t},...,pos_{T-1}\}$, where $pos_{t}\in \mathbb{R}^D$ is the sin-cos position embedding for frame $x_t$, defined as,  
\begin{equation*}
\label{eq:pos}
\centering
pos_{t}^{i}= \left\{  
\begin{array}{l}
\sin(w_k*t),i=2k\\
\cos(w_k*t),i=2k+1\\
\end{array}  
\right.,
w_k=\frac{1}{10000^{2k/D}}
\end{equation*}
where $D$ is the feature dimension, $pos_{t}^{i}$ is $i$th element of vector $pos_{t}$. Afterwards, we define the module input as,
\begin{equation}
\label{eq:recinput}
\centering
\left.
\begin{array}{ll}  
H_{0}(t)=&\{h_{0}, ..., \mathtt{[MASK]}_{t},..., h_{T-1} \}\\
&+\{pos_{0}, ,...,\mathtt{[MASK]}_{t},..., pos_{T-1}\}\\
\end{array}  
\right. 
\end{equation}
where $\mathtt{[MASK]}$ denotes that the corresponding token is masked at position $t$.

\begin{figure}[t!]
\centering
%
\includegraphics[width=0.95\linewidth]{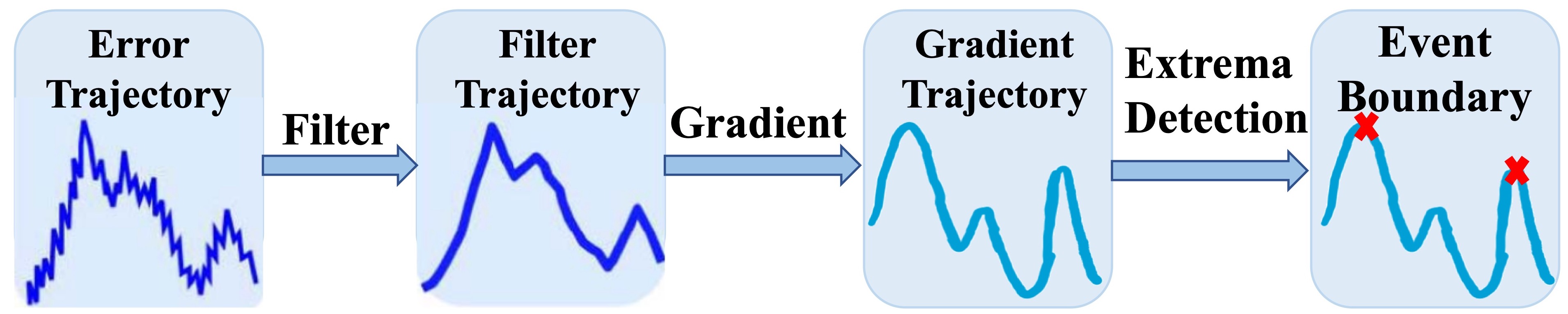}
\caption{The illustration of boundary detection from the reconstruction error trajectory.}

\label{fig:error}
\end{figure}



\noindent\textbf{Reconstruction Architecture.} To reconstruct the masked feature vector from $H_{0}(t)$, we modify the multi-head attention part of the transformer encoder. Specifically, we adopt 2 layers of Multi-head Self-Attention (MSA) and Multi-Layer Perceptron (MLP) blocks~\cite{vaswani2017attention} to process $H_{0}$ while randomly applying the mask $M(t)$ to the $t$-th feature embedding. The output of the $l$-th layer of the reconstruction module is defined as, 
\begin{equation}
\label{eq:transformer}
\centering
\left\{  
\begin{array}{l} 
H_{l}^{'} = \mathtt{MSA}(\mathtt{LN}(H_{l-1}),M(t)) + H_{l-1}\\
H_{l} = \mathtt{MLP}(\mathtt{LN}(H_{l}^{'})) + H_{l}^{'}\\
\end{array}  
\right. 
\end{equation}
where $H_{l-1}$ is the output of last $l-1$th layer, $\mathtt{LN}$ is the layer normalization~\cite{ba2016layer}.

\noindent\textbf{Reconstruction Objective.}  Previous work~\cite{devlin2018bert} formulates the mask reconstruction as a ``Masked Language Model''~(MLM) problem, 
which attempts to optimize the entropy of similarity between the reconstructed token and the original one~\cite{wang2019bert}. However, the embedded features of consecutive frames are very similar, which is not suitable for the entropy loss. Therefore, we directly treat it as an absolute feature reconstruction problem instead. To enforce the model to learn a local temporal relationship, we use the Mean-Squared-Error~(MSE) loss to measure the reconstructed feature:
\begin{equation}
\label{eq:reconstruction0}
\centering
\mathcal L_{R}=\mathbb{E}_{s\sim V,t\sim \{0,...,T-1\}}\|h_t^{'}-h_{t}\|_2^{2}
\end{equation}
where $V$ is the overall video dataset, $h_{t}^{'}$ is the reconstructed feature for the masked $t$-th frame, and $h_{t}$ is its original one. 



\subsection{Inference of CoSeg}
\label{sec:inference}
%
As the cognitive experiments shown~\cite{kurby2008segmentation}, a new event is triggered once people fail to anticipate it. Likewise, frames within an event are easier to be reconstructed than the boundary ones. This is reflected by the changes in the reconstruction errors. Hence, we apply the trained CoSeg system to detect event boundary using the reconstruction errors as illustrated in the right panel of Fig.~\ref{fig:framework}. 

During event boundary detection, CoSeg scans a video sequentially from its beginning to the end with a window of $T$ frames corresponding to a $T$-length video snippet or local context. Given a video snippet, we apply the frame feature reconstruction to its middle frame, then obtain a reconstruction error for each frame to form an error trajectory. Next, to detect boundaries from the error trajectory, we conduct the signal processing steps as follows (see Fig.~\ref{fig:error}).

1) \textbf{Filter Processing}: the FIR filter $S[t]=\sum_{k=-N}^{k=N}e_{k}E[t-k]$, where $e_{k}$ is the coefficient, is adopted to process the time-series signal $E$, which can reduce the noise and enhance the prominent features of the time series at the same time. Here we set $e_{k}=\frac{1}{2N+1}$ to smooth the error trajectory. 
%
%
2) \textbf{Gradient Calculation}: calculate gradient $G$ from $S$. 3) \textbf{Relative Extrema Detection}: The relative extrema detection~\cite{vemulapalli2012robust} is applied by picking up timestamp $t_{b}$ as a boundary by the following equation, 
\begin{equation}
\label{eq:relativeextrema}
\centering
\left\{  
\begin{array}{l} 
G[t_{b}]>G[t], t\in[t_{b}-r,t_{b}-1]\\
G[t_{b}]>G[t], t\in[t_{b}+1,t_{b}+r]
\end{array}  
\right. 
\end{equation}
where $r$ is the range for relative extrema detection.


\begin{table*}[]
\centering
\caption{Event boundary detection results on Kinetics-GEBD at different Rel.Dis. thresholds.}
\label{tab:gebd_threshold}
\begin{tabular}{llccccccccccc}
\toprule[1pt]
Supervision & Methods & 0.05 & 0.1&0.15&0.2&0.25&0.3&0.35&0.4&0.45&0.5&avg \\ \midrule[1pt]
\multirow{5}{*}{Supervised} & BMN~\cite{lin2019bmn} &0.186&0.204&0.213&0.220&0.226&0.230&0.233&0.237&0.239&0.241&0.223 \\
&BMN-StartEnd~\cite{lin2019bmn} &0.491&0.589&0.627&0.648&0.660&0.668&0.674&0.678&0.681&0.683&0.640  \\
 & TCN-TAPOS~\cite{lea2016segmental} & 0.464&0.560&0.602&0.628&0.645&0.659&0.669&0.676&0.682&0.687&0.627\\
 & TCN~\cite{lea2016segmental}  & 0.588&0.657&0.679&0.691&0.698&0.703&0.706&0.708&0.710&0.712&0.685\\
 & PC~\cite{shou2021generic} & 0.625&0.758&0.804&0.829&0.844&0.853&0.859&0.864&0.867&0.870&0.817 \\
\hline
\multirow{5}{*}{Unsupervised} & SceneDetect~\footnote{https://github.com/Breakthrough/PySceneDetect}&0.275&0.300&0.312&0.319&0.324&0.327&0.330&0.332&0.334&0.335&0.318\\
 & PA-Random~\cite{shou2021generic} & 0.336&0.435&0.484&0.512&0.529&0.541&0.548&0.554&0.558&0.561&0.506\\
 & PA~\cite{shou2021generic} & 0.396&0.488&0.520&0.534&0.544&0.550&0.555&0.558&0.561&0.564&0.527\\
 & \textbf{CoSeg} & \textbf{0.656} & \textbf{0.758} & \textbf{0.783}&\textbf{0.794}&\textbf{0.799}&\textbf{0.803}&\textbf{0.804}&\textbf{0.806}&\textbf{0.807}&\textbf{0.809}&\textbf{0.782}\\ \bottomrule[1pt]
\end{tabular}
\end{table*}

\begin{table}[]
\centering
\caption{The effectiveness of our Contrastive Temporal Feature Embedding (CTFE).}
\label{tab:contrastive}
\begin{tabular}{llccccc}
\toprule[1pt]
 Methods & MoF & IoU & Prec@5\%& Recall@5\%  & F1@5\% \\ \midrule[1pt]
\begin{tabular}[c]{@{}l@{}}ImageNet-VGG\\ Replacement \end{tabular} &49.2 & 47.4 & 41.4 & 35.4&38.2\\
\begin{tabular}[c]{@{}l@{}}ImageNet-ResNet\\ Replacement \end{tabular} & 50.7 & 47.3 &43.0&35.5&38.9 \\
Our CTFE CoSeg & \textbf{53.1} & \textbf{48.6} &\textbf{47.0}&\textbf{65.4}&\textbf{54.7}\\ 
 
  \bottomrule[1pt]
\end{tabular}
\end{table}

\begin{table}[]
\begin{center}
\caption{The effectiveness of Frame Feature Reconstruction.}
\label{tab:reclength}
\begin{tabular}{cccccccc}
\toprule
\begin{tabular}[c]{@{}l@{}}mask size $M$\\ Input len $T$\end{tabular} & $M$1 &$M$3 & \multicolumn{1}{l|}{$M$5} & $T$5 & $T$10 &  $T$20 &$T$30\\
\midrule
MoF & \textbf{53.1} & 53.0&\multicolumn{1}{l|}{52.8}  &51.6& \textbf{53.1} & 50.9& 51.1\\
IoU & \textbf{48.6} & 47.3&\multicolumn{1}{l|}{45.7}  &46.7& \textbf{48.6} & 47.5& 47.8\\
Prec@5\% & \textbf{47.0}  & 48.4 & \multicolumn{1}{l|}{47.6} & 45.8& \textbf{47.0} &47.2&47.3\\
Recall@5\% & \textbf{65.4} & 61.9 &\multicolumn{1}{l|}{63.4} & 63.3 &\textbf{65.4} &63.2&63.4\\
F1@5\% & \textbf{54.7}   & 54.3 & \multicolumn{1}{l|}{54.3} &  53.1 &\textbf{54.7} & 54.0& 54.1\\
\bottomrule
\end{tabular}
\end{center}
\end{table}

\section{Experiments}
\label{sec:exp}
\subsection{Implementation Details}
\label{sec:implement}
We adopt ResNet-18~\cite{he2016deep} as the backbone of the feature extractor, and modify the transformer encoder~\cite{vaswani2017attention} with 2 layers of 8 attention heads as the reconstruction architecture. In the contrastive feature embedding, the segment size $T$ is 10, and the temperature $\tau$ is set as 0.2 and the queue size $K$ of memory is 65536. The key encoder $g_\theta$ is the momentum update of query encoder with $g_\theta=\alpha*g_\theta+(1-\alpha)*f_\theta$, where $\alpha$ is set to 0.999. In temporal feature reconstruction, we mask 1 feature embedding to reconstruct it. The coefficient $\beta$ of overall loss in Eq.~(\ref{eq:overallloss}) is simply set to 1. We use the SGD~\cite{bottou2010large} optimizer to optimize the overall framework with the learning rate 0.002, weight decay 1e-4 and momentum 0.9. The training batch size $B$ is 16 and the number of video snippets from one video $c$ is 2, resulting in 32 video snippets in total. 
All the experiments are done in a server with 4 NVIDIA Tesla P40 GPUs. 

\begin{table*}[]
\centering
\caption{Transfer learning performance of CoSeg.} 
\label{tab:transfer}
\begin{tabular}{llccccc}
\toprule[1pt]

 Target & Source & MoF & IoU &Prec@5\% &Recall@5\% &F1@5\%  \\ \midrule[1pt]
 
\multirow{3}{*}{Breakfast}& Breakfast &\textbf{53.1} & \textbf{48.6}&\textbf{47.0}&	\textbf{65.4}	&\textbf{54.7}\\
 &INRIA& 51.9 & 48.1  &46.0&65.8&54.2\\
 &50salad & 51.9 & 48.2  & 46.3&66.4&54.5 \\ 
 \hline
 \multirow{3}{*}{INRIA}&INRIA& \textbf{47.9} & \textbf{42.4}& \textbf{46.7}	& \textbf{63.3} &\textbf{53.7}\\
 & Breakfast &45.7 & 42.2&45.2&66.6&53.9\\
 &50salad & 45.9 & 43.0&45.6&66.2&54.0 \\ 
 \hline
 \multirow{3}{*}{50salad}&50salad & \textbf{64.1} & \textbf{42.3} & \textbf{63.7}&\textbf{82.1}&\textbf{71.8} \\ 
  & Breakfast &60.7 & 42.0&56.9&90.0&69.7\\
 &INRIA& 62.9 & 41.9 & 57.1&90.7&70.0\\
 \bottomrule[1pt]
\end{tabular}
\end{table*}

\subsection{Datasets}
\label{sec:datasets}
Following the same protocol of  previous work~\cite{bhatnagar2017unsupervised,sener2018unsupervised,kukleva2019unsupervised,bhatnagar2017unsupervised,vidalmata2020joint}, we evaluated CoSeg on the following public datasets: Breakfast~\cite{kuehne2014language}, INRIA Instructional ~\cite{alayrac2016unsupervised}, 50Salads dataset~\cite{stein2013combining}, and the Generic Event Boundary Detection (GEBD) dataset\cite{kay2017kinetics}.

\noindent\textbf{Breakfast dataset} consists of 1,989 videos of 10 breakfast activities, where one person may perform different activities in different videos. The video qualities vary a lot due to occlusions and different viewpoints. 

\noindent\textbf{INRIA Instruction dataset} mainly contains 2-minute long videos collected from YouTube. There are about 47 sub-activities in this dataset. This dataset is very challenging for self-supervised event segmentation approaches, because it carries substantial number of ``background'' scenes, which generally do not share obvious common visual patterns. 

\noindent\textbf{50Salads dataset} collects multi-modal data from the cooking scenarios. In this dataset, the event segmentation is specified at different levels of granularity - high, low and eval. Following previous self-supervised works~\cite{lea2017temporal,lea2016segmental,aakur2019perceptual}, we use the ``eval'' granularity to train and evaluate our framework.

\textbf{Kinetics-GEBD} is the newest unconstrained video dataset specifically developed for the evaluation task of generic event boundary detection. It carries the largest number of event boundaries (e.g., 32x ActivityNet and 8x EPIC-Kitchens-100) from videos ``in the wild'', covering boundaries caused by action changes or generic event changes. Unlike other event segmentation datasets, such as Breakfast, INRIA Instruction and 50Salads, collected from a pre-defined taxonomy, Kinetics-GEBD is an open-vocabulary dataset. It consists of the train/val/test set with each having about 20K videos randomly sampled from the corresponding set of Kinectics-400 dataset~\cite{kay2017kinetics}, respectively.
The Kinectics-GEBD dataset labels the event boundaries from the following 5 different high-level causes: 1)Change of Subject; 2)Change of Object of Interaction; 3) Change of action; 4) Change in Environment; 5) Shot Change.


\begin{table}[]
\begin{center}
\caption{Sensitivity of error detection module.}
\label{tab:sensitivity}
\begin{tabular}{cccccc}
\toprule
\#Range & 50 & 60 & \textbf{70} & 80 & 90 \\
\midrule
MoF & 46.3 & 50.1 &\textbf{53.1}& 55.3 & 56.9\\
IoU & 47.6 & 48.5 & \textbf{48.6} & 48.2 & 47.5\\
Prec@5\% & 40.9 & 43.5&\textbf{47.0} & 48.3 & 49.6 \\
Recall@5\% &75.1 &71.4 &\textbf{65.4} & 62.5& 59.8\\
F1@5\% & 53.0 & 54.0 &\textbf{54.7} & 54.5 & 54.2\\
\bottomrule
\end{tabular}
\end{center}
\end{table}

\begin{table}[]
\centering
\caption{Segmentation results on the GEBD dataset.}
\label{tab:gebd}
\begin{tabular}{llccc}
\toprule[1pt]
Supervision & Methods &Prec@5\% & Rec@5\% & F1@5\% \\ \midrule[1pt]
\multirow{5}{*}{Supervised} & BMN~\cite{lin2019bmn} & 12.8 & 33.8 & 18.6\\
&BMN-StartEnd~\cite{lin2019bmn} & 39.6 & 64.8& 49.1  \\
 & TCN-TAPOS~\cite{lea2016segmental} & 51.8 & 42.0 &46.4  \\
 & TCN~\cite{lea2016segmental}  & 46.1 & 81.1&58.8 \\
 & PC~\cite{shou2021generic} & 62.4& 62.6&62.5  \\
\hline
\multirow{5}{*}{Unsupervised} & SceneDetect~\footnote{https://github.com/Breakthrough/PySceneDetect}&73.1&17.0&27.5\\
 & PA-Random~\cite{shou2021generic} & 6.8&17.7&9.9\\
 & PA~\cite{shou2021generic} & 26.5&22.2&24.2 \\
 & \textbf{CoSeg} & \textbf{60.4} & \textbf{71.8} & \textbf{65.6}\\ \bottomrule[1pt]
\end{tabular}
\end{table}

\subsection{Evaluation Metrics}
\label{sec:eval_metric}
As aforementioned, our goal is to detect generic event boundaries from videos ``in the wild'', where a boundary is defined as a timestamp or a short range represented by its middle timestamp. To evaluate the detection algorithm, the discrepancy between the detected boundary and the ground truth boundary is the most important measurement for us. In this work, we adopt the Precision/Recall and F1 score as our primary evaluation metric. Specifically, our evaluation formula is similar to that of the Generic Event Boundary Detection challenge held with CVPR 2021~\cite{shou2021generic}. Just like the Intersection-over-Union~(IoU) measurement, the Relative Distance~(Rel.Dis.) is calculated as the error between the detected and ground truth boundary, divided by the length of the corresponding video instance. Given a fixed threshold of Rel.Dis (e.g., 5\% in our evaluation), the Precision/Recall and F1 score are computed by determining if a detection is correct or not within the threshold. Notice the F1 score is also widely used in supervised event segmentation tasks~\cite{lea2016segmental,lea2017temporal}. These evaluation metrics can be evaluated in matching-free style, which can provide more direct and fair comparison across different methods.

Additionally, we also try some additional evaluation metrics including the Mean Over Frames~(MoF) and Intersection Over Union~(IoU), which are widely used to evaluate the (weakly) supervised or clustering-based unsupervised \textbf{non-generic} event segmentation works~\cite{bhatnagar2017unsupervised,sener2018unsupervised,kukleva2019unsupervised,bhatnagar2017unsupervised,vidalmata2020joint}. To calculate these metrics, the Hungarian matching algorithm needs to be applied to build a one-to-one match between the predicted segments and the ground truth event segments. This matching is conducted for a set of pre-defined event categories. However,  since our work targets at generic event boundary regardless of its event category, we are unable to perform the global matching like ~\cite{sener2018unsupervised}. Instead, we apply the matching at video-level. In other words, as we do not assign an event label or pseudo label (clustering based approaches) to the segments, our matching varies by video. Therefore, we should be aware that we can not directly compare our approach to previous works by means of MoF and IoU. It is just for a reference to list the results in terms of these metrics.

\begin{table*}[]

\centering
\caption{Segmentation results on the Breakfast dataset. }
\label{tab:breakfast}
\begin{tabular}{llcccccc}
\toprule[1pt]
Supervision & Methods & MoF & IoU & Prec@5\%& Recall@5\%  & F1@5\%  \\ \midrule[1pt]
 
\multirow{5}{*}{Unsupervised} & KNN+GMM~\cite{sener2018unsupervised} & 34.6 & 47.1&-&-&-\\
 & CTE-MLP~\cite{kukleva2019unsupervised} & 41.8 & - &32.5 &30.3 & 31.3\\
 & U-Net~\cite{vidalmata2020joint} & 42.7 & 12.8&-&-&29.9\\
 & LSTM+AL~\cite{aakur2019perceptual} & 42.9* & 46.9* &-&-&-\\
 & \textbf{CoSeg} & \textbf{53.1*} & \textbf{48.6*}&\textbf{47.0}&	\textbf{65.4}	&\textbf{54.7}\\ \bottomrule[1pt]
 
\end{tabular}
\begin{tablenotes}
      \small
      \item * indicates the metric calculation without global matching
\end{tablenotes}
\end{table*}


\begin{table*}[]
\centering
\caption{Segmentation results on the INRIA dataset.}
\label{tab:YTI}
\begin{tabular}{llccccc}
\toprule[1pt]
Supervision & Methods & MoF & IoU &Prec@5\% & Recall@5\% & F1@5\% \\ \midrule[1pt]

\multirow{4}{*}{Unsupervised} & KNN+GMM~\cite{sener2018unsupervised} & 27.0 & -  &-&-&-\\
 & CTE-MLP~\cite{kukleva2019unsupervised} & 39.0 & 9.6 &61.6&18.5&28.4\\
 & U-Net~\cite{vidalmata2020joint} & 39.1 & 9.4&-&-&29.9 \\
 & \textbf{CoSeg} & \textbf{47.9*} & \textbf{42.4*}& \textbf{46.7}	& \textbf{63.3} &\textbf{53.7}\\ \bottomrule[1pt]
\end{tabular}
\end{table*}

\begin{table*}[]
\centering
\caption{Segmentation results on the 50 Salads dataset. }
\label{tab:salad}
\begin{tabular}{llccccc}
\toprule[1pt]
Supervision & Methods & MoF & IoU  & Prec@5\% & Recall@5\% &F1@5\% \\ \midrule[1pt]
\multirow{5}{*}{Unsupervised} & 
 CTE-MLP~\cite{kukleva2019unsupervised} & 35.5 & - &74.9&36.5&49 \\
 & LSTM + KNN~\cite{bhatnagar2017unsupervised} & 54.0 & - &-&-&-\\
 & LSTM+AL~\cite{aakur2019perceptual} & 60.6* & -&-&-&-\\
 & U-net~\cite{vidalmata2020joint} & -& -&-&-&56.9\\
 & \textbf{CoSeg} & \textbf{64.1*} & \textbf{42.3*} & \textbf{63.7}&\textbf{82.1}&\textbf{71.8}\\ \bottomrule[1pt]
\end{tabular}
\end{table*}

 
Let us assume there are $U$ segments in the ground truth, and consider a matched pair ($Y$,$Z$) where $Y$ is a predicted segment and $Z$ is a ground truth. Further, let $Y\cap Z$ denote the intersection of two segments and $Y\cup Z$ represent their union. Then the metrics Mean Over Frames~(MoF), Intersection Over Union~(IoU) can be defined as, 
\begin{equation}
\label{eq:metric}
\centering
\left\{
\begin{array}{l}
 MoF= \frac{\sum_{k=1}^{k=U}Y_{k}\cap Z_{k}}{\sum_{k}^{k=U}Z_{k}}\\
IoU= \frac{1}{U}\sum_{k=1}^{k=U}\frac{Y_{k}\cap Z_{k}}{Y_{k}\cup Z_{k}} \\
\end{array}
\right.
\end{equation}


\subsection{Performance Analysis on CoSeg}
\label{sec:analysis}
\noindent\textbf{Performance at different Rel.Dis. thresholds.} The primary evaluation metric for generic event detection is F1 scores (calculated from Precision/Recall) at a pre-defined Rel.Dis. threshold by users. In general, a smaller threshold implies higher criterion to determine whether a boundary detection is correct. In other words, the error between the detected boundary and ground truth is smaller. Table~\ref{tab:gebd_threshold} displays the event boundary detection F1 scores on Kinectics-GEBD dataset given various Rel.Dis. thresholds. When the criterion is getting easier (i.e., higher Rel.Dis. threshold), the F1 score is increasing. We also notice that when the threshold is over 0.2, the change of F1 score is minor. In the rest of our experiments, we only report the Precision/Recall/F1 score at 5\%, which is the strictest criteria to calculate F1.      

\noindent\textbf{Effectiveness of the CTFE module.} In this group of experiments, we want to verify the effectiveness of the proposed Contrastive Temporal Feature Embedding (CTFE, see Section~\ref{sec:contrastive}) in our CoSeg system. To this end, we replaced the self-trained CTFE module with an ImageNet pre-trained VGG~\cite{simonyan2014very} (ImageNet-VGG) or ResNet~\cite{he2016deep} (ImageNet-RN) encoder. All relevant experiments were conducted on Breakfrast dataset with default settings. Table~\ref{tab:contrastive} illustrates their performance comparisons in terms of various evaluation metrics. It clearly shows that the CTFE approach can outperform both ImageNet pre-trained models by a large margin in all evaluation metrics, such as by 2\%-3\% in MoF, 1\%-2\% in IoU, 16\%-17\% in F1@5\%. These improvements actually confirm our conjecture that our CTFE can generate better feature representation for event segmentation, thanks to its capability of capturing the underlying temporal difference between frames of the inter-event and intra-event. In addition, please note that unlike the ImageNet pre-trained models, our feature embedding does not require to be trained with external datasets, which makes the end-to-end CoSeg framework more flexible.


\noindent\textbf{Effectiveness of the FFR module.} We further conducted a set of experiments to verify the effects of various settings for our Frame Feature Reconstruction (FFR) module (see Section~\ref{sec:reconstruction}). The experimental results on Breakfast dataset are shown in Table~\ref{tab:reclength}. The three columns (MX) in the left panel of the table display the results in terms of various metrics achieved by different reconstruction size~(i.e., mask size, the number of frames to be reconstructed, 1/3/5 in our experiments) when fixing the window size $T=10$. As we can see, the mask size of 1 (M1 column) achieves the best performance. It seems simultaneously reconstructing more than one frame may decrease the event boundary detection accuracy. Meanwhile, we also examined how the value of video snippet length $T$ (i.e., the size of local context or window) affects the boundary detection. In terms of all evaluation metrics, $T=10$ consistently outperform other snippet length/window size. It is worth noting that adding longer window size (local supportive context) does not necessary improve the detection accuracy. This observation is consistent with the fact that humans can quickly detect a new event without knowing much context.  

\noindent\textbf{Generalizability of CoSeg framework.} Although the long-term memory is built over various past experiences, humans are able to detect event boundaries under a completely new situation. This is the capability of generalization.
Likewise, can we train the CoSeg with different datasets and test it on an unseen dataset? To answer this, we designed a set of cross-dataset experiments, in which we trained the CoSeg model from the source dataset and apply the model for event segmentation on the target dataset. All results are shown in Table~\ref{tab:transfer}. We notice that our CoSeg has a strong capability of generalization across datasets. In other words, the CoSeg model trained on different datasets can achieve similar performance. In particular, the INRIA Instructional dataset looks very different from the Breakfast and 50Salads dataset, but the performance of cross-dataset transfer is still promising. These experiments tell us that the generalizability of CoSeg enable our model to detect more generic event boundary from videos ``in the wild''. 


\noindent\textbf{Sensitivity to error detection hyper-parameter.} To investigate the influence of hyper-parameters of error detection, we examine the performance on the Breakfast dataset with different ``range'' parameters $r$~(see Eq.~(\ref{eq:relativeextrema})) to detect the extrema as an event boundary. As shown in Table~\ref{tab:sensitivity}, our performance is stable under different hyper-parameters for detecting the event boundaries. Here smaller $r$ leads to smaller MoF. That indicates a smaller $r$ has more segments and shows less preference to the dominant classes, while higher $r$ demonstrates preferences to dominant classes and results fewer segments. But different $r$ always yield higher values on different metrics. To conclude, our extraordinary performance is not dependent on the error detection hyper-parameters.

\subsection{Performance comparisons to STOA approaches}
\label{sec:performance}
The task of video event segmentation has been developed for many years, which results in various benchmark datasets and evaluation metrics. In this subsection, we will compare the performance of our CoSeg with other STOA approaches on different benchmark datasets with 5 evaluation metrics.

\noindent\textbf{Kinetics-GEBD Dataset.} As aforementioned, Kinectics-GEBD is the most challenging and largest benchmark for generic event boundary detection. We first want to demonstrate how well our CoSeg performs on this dataset as comparing with other approaches. We trained the models on the \emph{train} set and tested them on \emph{val} set ( The \emph{test} set is not publically available). Table \ref{tab:gebd} presents the F1 scores of event boundary detection at different Rel.Dis thresholds from our CoSeg system as well as other SOTA approaches. 
As we can see, our CoSeg outperforms all unsupervised approaches by a large margin. In terms of average F1 scores at all thresholds, our F1 score is about 26\% better than the second best. Our approach is even better than most of supervised approaches as reported in ~\cite{shou2021generic}, and is comparable to the best supervised approach PC ~\cite{shou2021generic} (Only about 4\% lower in average F1 scores than PC method ).

\noindent\textbf{Breakfast Actions Dataset.} This dataset was prepared for the evaluation of previous supervised and cluster-based unsupervised event segmentation, which deals with a set of pre-defined event categories rather than generic events. The MoF and IoU metric are widely used to evaluate the pre-defined event segmentation algorithms. To compare with previous results, we also calculated both metrics on this dataset. However, as discussed in ~\ref{sec:eval_metric}, we are unable to directly compare our method to previous works in MoF and IoU. To make a fair comparison, we also calculated the Precision/Recall/F1 score @5\% threshold for this dataset. 

Table~\ref{tab:breakfast} displays all results in all 5 metrics. Most previous work do not do not report the Precision/Recall/F1 scores, and we are unable to acquire their source code. Hence, we will leave these metric values empty for these work. It is obvious that CoSeg outperforms all previous work in all evaluation metrics. The comparison between our work and LSTM+AL is meaningful in terms of MoF and IoU, as both do not have global matching (marked with * in the table). Higher IoU value indicates CoSeg does not suffer from the over-segmentation problem even with a higher MoF value. In terms of F1@5\% score, our work is significantly better (over 20\% higher) than the latest U-Net and CTE-MLP work, which is very promising. It further proves the effectiveness of our end-to-end self-supervised framework. 

\noindent\textbf{INRIA Instructional Video Dataset.} INRIA dataset is more challenging than Breakfast dataset due to its wide spread ``Background'' segments. We repeated the evaluation of Breakfast onto INRIA dataset. The comparisons to other approaches are listed in Table.~\ref{tab:YTI}. Most previous approaches are biased due to their low IoU scores, which suggests those methods prefer dominant classes instead of finding the correct segmentation. In contrast, our CoSeg achieves the best performance in all metrics, and it can keep a good balance between dominant and underrepresented events. 
The superior results have shown our self-supervised framework integrated with long-term and online memory can deal with the ``Background'' segments and detect the proper segmentation with much less under-segmentation. Most importantly, we yield 63.3\% recall of ground truth boundary while CTE-MLP only recalled 18.5\%, which means large percents of boundaries are missed.


\noindent\textbf{50-Salads Dataset.} 50-Salads is a multi-modality dataset, but we only used its RGB modality in our experiments. Following the experiments on previous two datasets, we illustrate the experimental results on 50-Salads in Table~\ref{tab:salad}. As we can see, CoSeg can still significantly outperform the recent U-Net and CTE-MLP in F1 score @5\%. Under the same MoF calculation, our approach can still beat LSTM+AL approach.


In all 4 datasets, our framework greatly improves the performance without using annotated data or external data. Meanwhile, compared to previous unsupervised methods, we achieved higher MoF, IoU at the same time. It suggests our method alleviates the overfitting problem which widely existed in previous methods. In addition, a much higher F1@5\% also suggests that segmentation results of CoSeg have less over- or under- segmentation instances.  




\subsection{Qualitative Evaluation}
\label{sec:visualize}

\begin{figure*}[!htb]
\centering
%
\includegraphics[width=\linewidth]{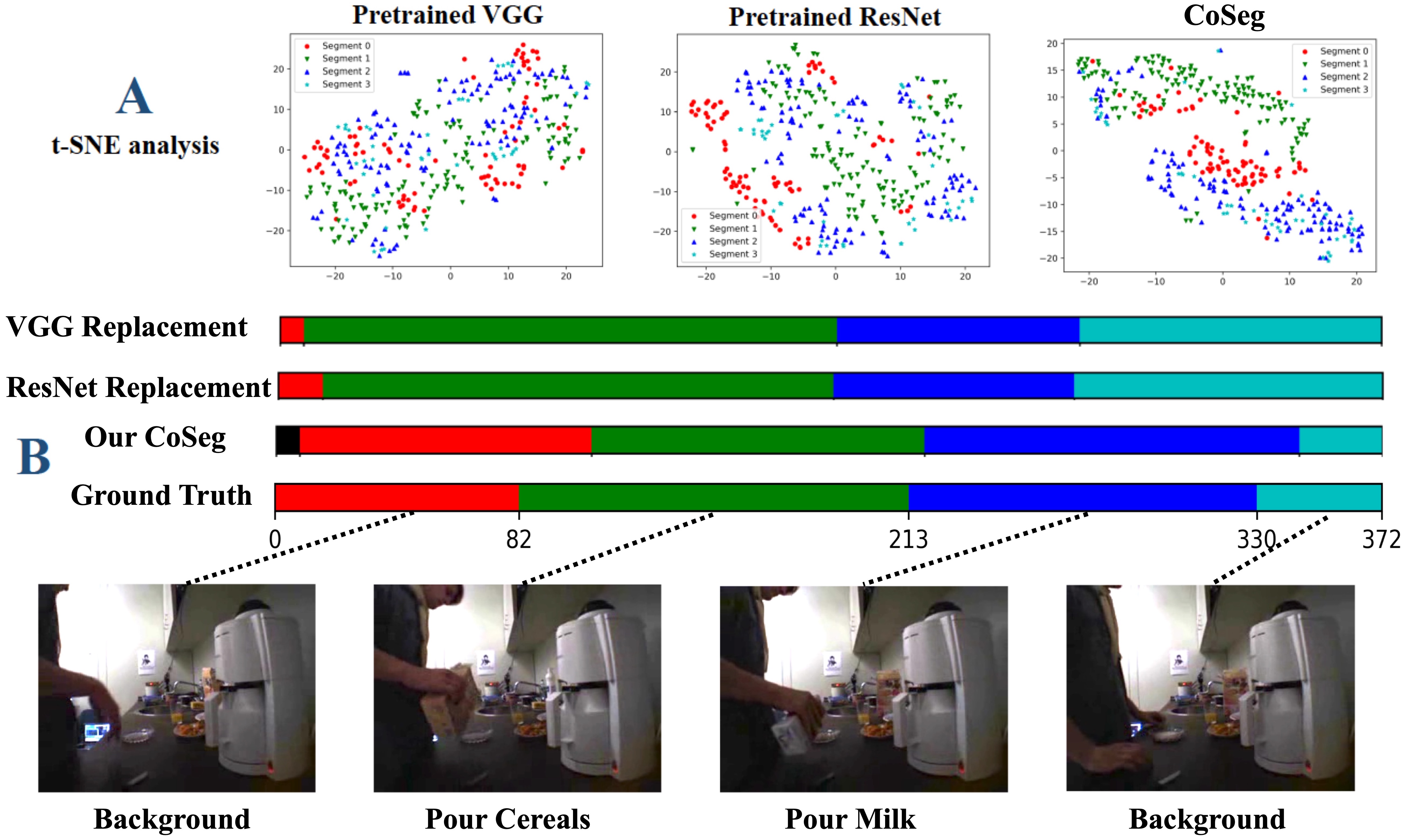}
\caption{Qualitative analysis of CoSeg on Breakfast dataset~\cite{kuehne2014language}. This example is from P33\_cam03\_cereals video. (A) Panel A compares the t-SNE~\cite{maaten2008visualizing} of the extracted visual features by the ImageNet-pretrained VGG and ResNet network, as well as our CoSeg. Self-supervised visual features show a better clustering in terms of the ``Pour cereals" and ``Pour milk" events. It can even group the frames without meaningful actions with the knowledge learned from the entire dataset (shown in red color). (B) Our segments agree well with the ground truth labels, even though the overall background and visual conditions remain the same, which increases the challenge for detecting events. Other models fail to segment segments 0 and 1 due to their weak visual representations.}
\vspace*{-\baselineskip}
\label{fig:qualitative}
\end{figure*}

In this subsection, we will demonstrate and analyze some qualitative results by visualizing the event segmentation results of our CoSeg on the Breakfast Dataset and 50 Salads Dataset.

Fig.~\ref{fig:qualitative} illustrates the visualization comparisons of event segmentation results for a video (P33\_cam03\_cereals) from Breakfast dataset using different model. In the upper panel A, it displays three t-SNE visualization results of visual features learned by VGG, ResNet and our CoSeg network, respectively, where each color represents one event segment/cluster. As we can see from the t-SNE visualization, CoSeg demonstrates much better feature aggregation than other pre-trained networks, e.g., the green and blue features corresponding to event ``pour cereals'' (segment 1) and ``pour milk'' (segment 2). It visually verifies that similar visual features can be more easier to be clustered under the self-supervised mechanism. In panel B, it further illustrates the event segmentation results of our CoSeg and its variants (the replacement with ImageNet-VGG or ImageNet-ResNet). We can observe that the segmentation of CoSeg clearly shows better matching to the ground truth. Although the features extracted from segment 2 and 3 are intertwined with each other, our CoSeg can still detect the event boundary between segment 2 and 3, thanks to the advance of Frame Feature Reconstruction module.  

\begin{figure*}[!htb]
\centering
\includegraphics[width= \linewidth]{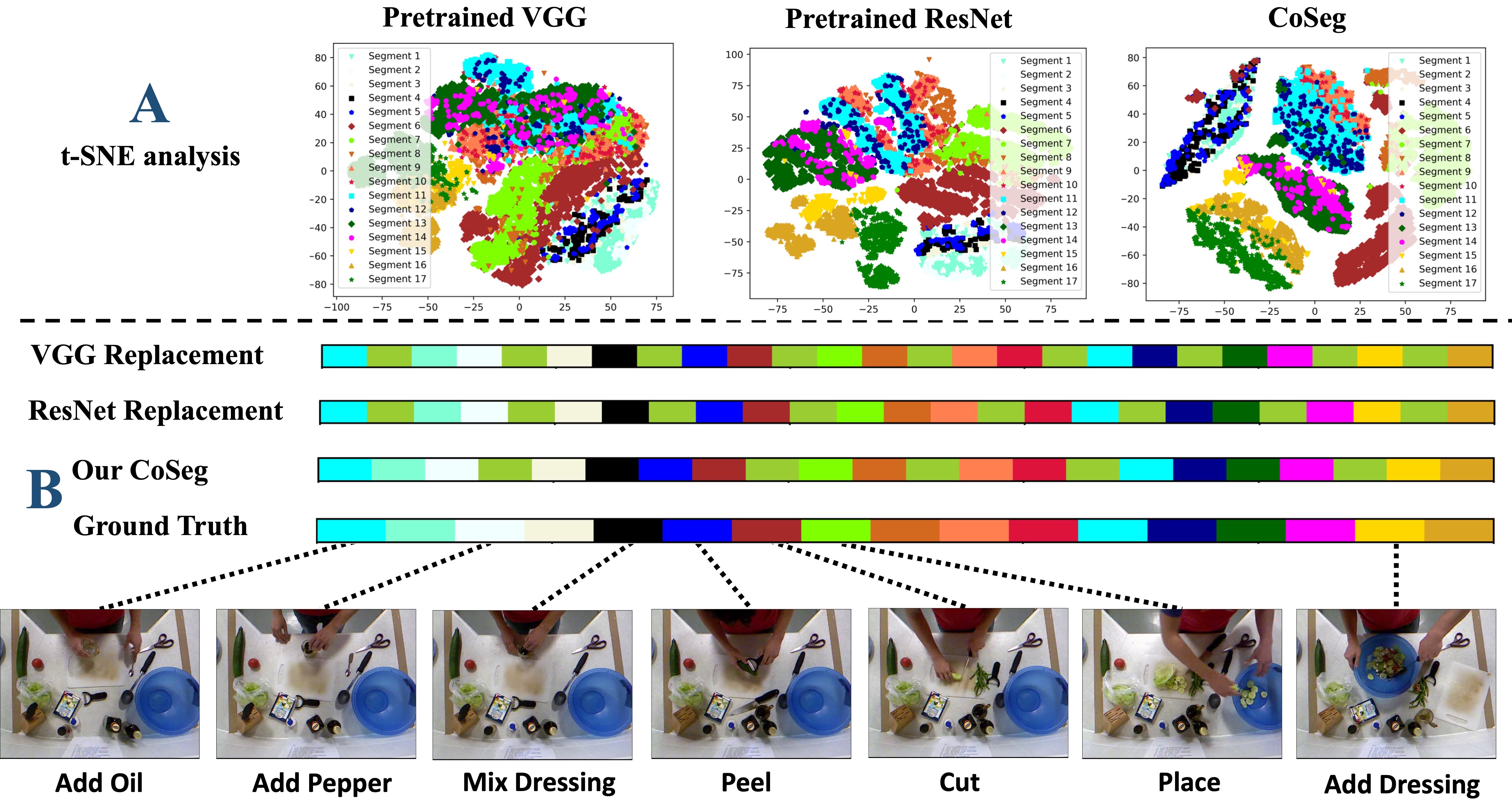}
\caption{Qualitative analysis of CoSeg on 50 Salad Dataset. This example is from rgb-22-1 video of the 50 Salads dataset
and is a good example to show over-segmentation problem for long videos. \textbf{A}. The  t-SNE~\cite{maaten2008visualizing} of the extracted visual features by the ImageNet-pretrained VGG and ResNet network, as well as temporal contrastive feature embedding of CoSeg.  \textbf{B}. Temporal segmentation time lines of different approaches.(non-match segments are shown in lightgreen). }
\label{fig:qualitativesalad}
\end{figure*}

Fig.~\ref{fig:qualitativesalad} shows another visualization example from 50 Salads Dataset. We can have similar observations as previous one. Although this video has very long duration and carries many events, our CoSeg system can still detect the event boundary accurately. In addition, the t-SNE visualization further demonstrates that features trained under our contrastive embedding module can achieve better aggregation than other alternative embedding solutions. Fewer segments detected by CoSeg suggest that our boundary detection is more conservative and more precise, which does not suffer over-segmentation problem for complicated scenario existed in long video.

\section{Conclusion}
\label{sec:conclusion}

In this paper, we propose a self-supervised framework, called CoSeg, for generic event boundary detection or event segmentation. The system design is inspired by some recent cognitive research on how humans detect new events. First of all, we invent a Contrastive Temporal Feature Embedding strategy to acquire semantic meaningful feature representation, which mimics the function of human's long-term memory. Secondly, we formulate the event segmentation problem as a Frame Feature Reconstruction task, which is similar to the behavior of human predicting new events. We thoroughly evaluate our framework on four datasets. The experiments demonstrate that our framework can outperform other solutions by a big margin. Furthermore, compared to previous works, no prior information is needed for CoSeg so that it can be readily applied to new datasets with competitive performance. Most notably, CoSeg has achieved the state-of-the-art results on GEBD dataset, which clearly prove its generalizability on general event boundary detection tasks. 


\ifCLASSOPTIONcompsoc
  \section*{Acknowledgments}
\else
  \section*{Acknowledgment}
\fi
 The author would like to thanks Rosaura's kind help in evaluating their method Unet~\cite{vidalmata2020joint} with our new evaluation metrics. The author also would like to thanks Yuanyuan Zhang for her help in the figure revising.

{\small
\bibliographystyle{IEEEtran}
\bibliography{egbib}
}

\begin{IEEEbiography}[{\includegraphics[width=1in,height=1.25in,clip,keepaspectratio]{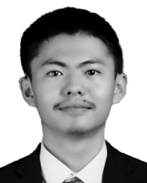}}]{Xiao Wang}
	Xiao Wang received his B.S. degree from Department of Computer Science in Xi’an Jiaotong University, Xi’an, China in 2018. He is currently pursuing his Ph.D. degree in the Department of Computer Science at Purdue University, West Lafayette, IN, USA. His research interests include representation learning, deep learning, computer vision, bioinformatics and intelligent systems. 
\end{IEEEbiography}

\begin{IEEEbiography}[{\includegraphics[width=1in,height=1.25in,clip,keepaspectratio]{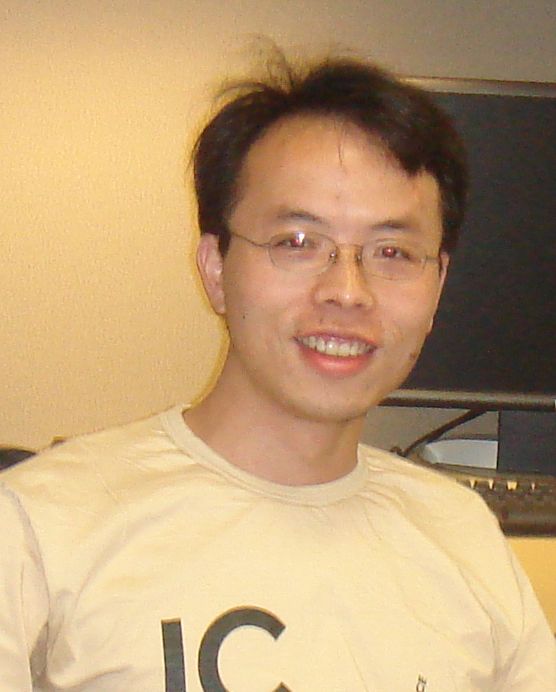}}]{Jingen Liu}
Dr. Jingen Liu is a Computer Vision Researcher at JD AI Research (JD.com Silicon Valley Research Center). Before joining JD.com, he was a Computer Scientist at SRI International from 2011 to 2018, and a research fellow at University of Michigan, Ann Arbor from 2010 to 2011. He received his Ph.D. degree from the Center for Research in Computer Vision at University of Central Florida in 2009. His research is focused on computer vision, multimedia processing, medical image analysis and machine learning. His expertise lies in video analysis, human action recognition, multimedia event detection, semantic segmentation, and deep learning. He has published 50+ papers on prestigious AI conferences, such as CVPR, ICCV, ECCV, AAAI, and ACM Multimedia. As an outstanding young scientist, he was invited by NAE to present his work on video content analysis in the Japan-American Frontiers of Engineering Symposium 2012. He is the associate editors for international journals including IEEE Transaction on CSVT, Pattern Recognition, and IAPR Machine Vision \& Applications, and an area chair of IEEE WACV 2015 \& 2016. He co-chaired the THUMOS series, the large-scale action classification \& detection evaluation workshops (2013 \& 2014). 
\end{IEEEbiography}

\begin{IEEEbiography}[{\includegraphics[width=1in,height=1.25in,clip,keepaspectratio]{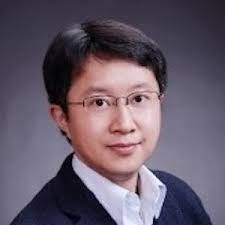}}]{Tao Mei} Tao Mei is a technical vice president with JD.com and the deputy managing director of JD AI Research, where he also serves as the director of the Computer Vision and Multimedia Lab. Prior to joining JD.com in 2018, he was a senior research manager with
Microsoft Research Asia in Beijing, China, where he contributed 20 inventions and technologies to Microsoft’s products and services. He has authored or co-authored more than 200 publications (with 11 best paper awards) and holds 20 US granted patents. He was or has been an editorial board member of the IEEE Transactions on Image Processing, the IEEE Transactions on Circuits and Systems for Video Technology, the IEEE Transactions on Multimedia, the ACM Transactions on Multimedia Computing, Communications, and Applications, and the ACM Transactions on Intelligent Systems and Technology. He is the general co-chair of IEEE ICME 2019, the program co-chair of the ACM Multimedia 2018, IEEE ICME 2015 and IEEE MMSP 2015. He was elected as a fellow of IAPR and a distinguished scientist of ACM in 2016, for his contributions to large-scale multimedia analysis and applications. He is also a distinguished industry speaker of the IEEE Signal Processing Society. He is a fellow of the IEEE.
\end{IEEEbiography}

\begin{IEEEbiography}[{\includegraphics[width=1in,height=1.25in,clip,keepaspectratio]{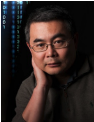}}]{Jiebo Luo}
	Jiebo Luo (S93, M96, SM99, F09) is a Professor of Computer Science at the University of Rochester which he joined in 2011 after a prolific career of fifteen years at Kodak Research Laboratories. He has authored nearly 500 technical papers and holds over 90 U.S. patents. His research interests include computer vision, NLP, machine learning, data mining, computational social science, and digital health. He has been involved in numerous technical conferences, including serving as a program co-chair of ACM Multimedia 2010, IEEE CVPR 2012, ACM ICMR 2016, and IEEE ICIP 2017, as well as a general co-chair of ACM Multimedia 2018. He has served on the editorial boards of the IEEE Transactions on Pattern Analysis and Machine Intelligence (TPAMI), IEEE Transactions on Multimedia (TMM), IEEE Transactions on Circuits and Systems for Video Technology (TCSVT), IEEE Transactions on Big Data (TBD), ACM Transactions on Intelligent Systems and Technology (TIST), Pattern Recognition, Knowledge and Information Systems (KAIS), Machine Vision and Applications, and Journal of Electronic Imaging. He is the current Editor-in-Chief of the IEEE Transactions on Multimedia. Professor Luo is also a Fellow of ACM, AAAI, SPIE, and IAPR.
\end{IEEEbiography}

\end{document}